# Natural Language in Requirements Engineering for Structure Inference - An Integrative Review

Maximilian Vierlboeck, M.Sc., Carlo Lipizzi, Ph.D, and Roshanak R. Nilchiani, Ph.D.

*Abstract —* **The automatic extraction of structure from text can be difficult for machines. Yet, the elicitation of this information can provide many benefits and opportunities for various applications. Benefits have also been identified for the area of Requirements Engineering. To evaluate what work has been done and is currently available, the paper at hand provides an integrative review regarding Natural Language Processing (NLP) tools for Requirements Engineering. This assessment was conducted to provide a foundation for future work as well as deduce insights from the stats quo. To conduct the review, the history of Requirements Engineering and NLP are described as well as an evaluation of over 136 NLP tools. To assess these tools, a set of criteria was defined. The results are that currently no open source approach exists that allows for the direct/primary extraction of information structure and even closed source solutions show limitations such as supervision or input limitations, which eliminates the possibility for fully automatic and universal application. As a results, the authors deduce that the current approaches are not applicable and a different methodology is necessary. An approach that allows for individual management of the algorithm, knowledge base, and text corpus is a possibility being pursued.**

*IndexTerms -* **natural language processing, requirements engineering, systems engineering, product development**

## I. INTRODUCTION, SITUATION, AND PROBLEM

The extraction of structure and schematic information from a body of text can seem like a trivial task. Furthermore, drawing models, graphs, and networks as a representation of language and thought can be an efficient way to communicate and bring certain insights to light. As straight forward as this task might seem though, automating or having it done even just in part by a machine can be difficult due to a multitude of factors, such as subjectivity [1], ambiguity [2], and domain specific circumstances. These factors can make universal application of tools challenging or require limitations to be defined and imposed.

An area where structural analysis is of importance is Requirements Engineering (RE). Herein, when the requirements to fulfill by a product or system are defined, potentially extensive corpora of text are created that are interrelated, yet not necessarily structured accordingly. Therefore, eliciting the structure based on the language and semantics can allow for various analysis possibilities of the requirement body. One possible result of such an analysis could be a systemigram or network showing the different connections within the specifications. Due to these benefits and potential use cases the research at hand addresses the possibilities and opportunities within the field of Requirements Engineerings to contribute to the understanding and efficiency.

A popular approach to extract information and or structure from text or speech is Natural Language processing (NLP). Despite different and sometimes controversial definitions of NLP [3], it is in its core the attempt to process natural language with computer tools and methods that are supposed to allow a human-like linguistic analysis and potential manipulation of text or speech [3-5]. As such, processing and extracting structure from text is one possible implementation of NLP. Yet, the possibilities and research directions of NLP are manifold and as such, the aforementioned extraction of structure is just one of the problems that can also be approached in different ways. This diversity lead to the development of numerous tools over time that could be used to extract certain types of structure from source text or speech. Yet, the actual applicability and usefulness of existing tools has to be evaluated due to various limitations, for example. This can make not only the research of existing tools difficult since most have to be carefully assessed for their criteria, but also complicates the continuation of research because of the crowdedness and diversity of the space. This is why the work at hand attempts to provide insight and address the complicatedness in form of a structured literature review and analysis of the space regarding the computer-aided and or automatic extraction of semantic structure from natural language which is also considered a current challenge of NLP [6]. Furthermore, this analysis and review shall enable and streamline subsequent research.

Based on the problem and opportunities described above, the paper at hand has been divided into six sections. This first section intrudes the situation and topic to address. The second section outlines the applied methodology and utilized tools to conduct the review and analysis. Following the methodology, the third section describes the history and literature that was considered to outline the frame and set the foundation for the analysis. Said analysis is subsequently described in the fourth section including the description of all assumptions and limitations. The second to last and fifth section discusses the findings before the sixth section finally concludes the presented work with the inclusion of a summary and the limitations.

II. METHODOLOGY AND APPROACH

In order to conduct the described review and analysis in a valid and scientific way that allows for relation to existing knowledge, a methodology that allows for tracing and reproducibility is crucial. While the often described as "gold standard" of reviews, the systematic literature review [7], is a possibility, it is not usable in its full form due to the fact that the scientific area is diverse and shows a multitude of directions and options as aforementioned, which would create an almost insurmountable body of literature to assess. Thus, a semi-structured review, following the process of the development over time would be the other systematic option. This possibility has been decided to be not suitable for the task at hand though, despite the consideration of different concepts within the NLP space, due to the fact that the objective of the presented work is not the consideration of individual interpretations and approaches, but the comprehensive assessment of the space to gain insight into the overall setup and thus enable the analysis. As a result, an integrative review has been chosen since it allows to evaluate and conflate existing literature regarding a matter and or popular topic to allow the derivation of different frameworks and view points [7, 8]. Furthermore, integrative reviews have been applied in similar fashion to other contexts [8, 9].

The integrative procedure of the literature review was conducted as follows and is depicted in Figure 1 below. In the first step, (sections one through three) the setting and frame were defined in addition to the selection of the sources to consider. This selection was approached by first assessing an overview of the field and then outlining the potential sources to evaluate based on those insights. Herein, recent works were discovered that could be utilized due to their peer reviewed nature to serve as a starting point that additional sources could be derived from. Overall, all sources found that address the topics at hand (see section three for further details) are included.

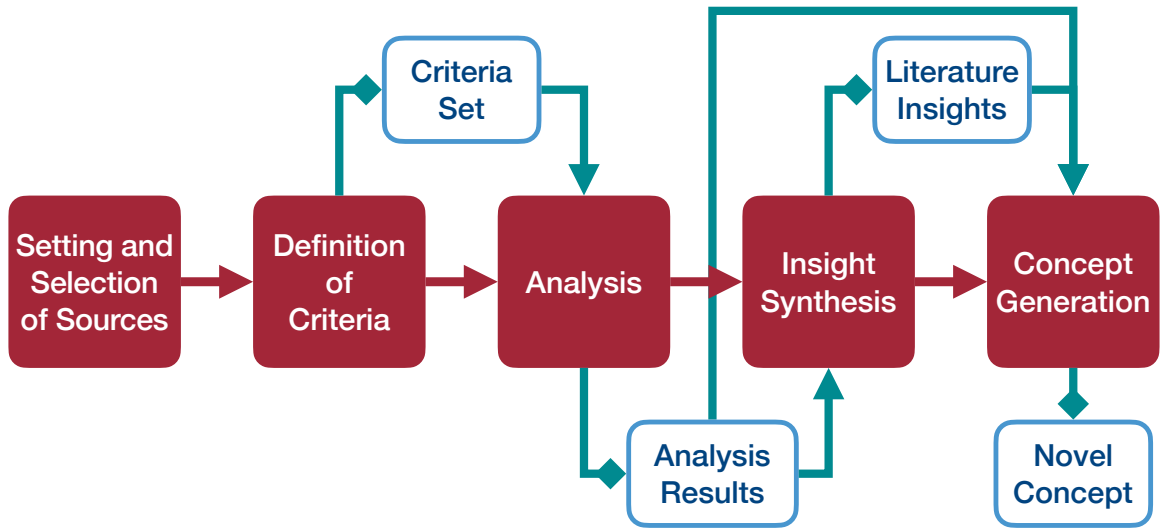

*Fig. 1 - Research Process and Steps*

With the sources at hand, the criteria were defined according to which the analysis was supposed to be conducted. The set of criteria was derived from the objectives at hand, namely RE and structure elicitation via NLP. The exact selection and definition of each criterion including the reasoning is outlined in the first sub-section of section four. The criteria enabled the analysis which is the core of the integral analysis and allows for the definition of an overview of the space and scientific field, which forms the second sub-section of section four. From the analysis, the actual insights could be derived and formulated to enable for the design of a new concept based on the potential shortcomings of the current situation. Due to the extent of the review and effort, the actual design and concept generation will be subject of a future publication.

With the methodology outlined above, the review was conducted in a structured way according to the guidelines and recommendations of a integrative review. The deduced conclusions and insights can be seen as valid as the quality of the sources, which are described and explained in the next section.

### III. HISTORY AND LITERATURE

Due to integrative nature of the presented review and analysis, a complete assessment of the entire literature available regarding NLP and or RE is not expedient. Yet, in order to outline the frame and setting, an overview shall be presented first before the current situation and sources can be addressed. As such, the following two sub-sections take a look at the history of NLP first before bringing it together with RE to assess the current state of both topics as a combinatory field and research direction.

*Natural Language Processing History and Progress*

Looking at NLP from a general perspective, three domains emerge as the main drivers: *Linguistics*, *Computer Science*, and *Psychology* [3]. The first field, Linguistics, is concerned with the structural and formal aspects of language; the second one, Computer Science, focuses on the processing and structuring of data; and the last one, Psychology, contributes the insights into cognitive processes and psychological models of language. As a result, two directions exist in NLP: language processing and language generation. The language processing on one hand analyses existing text/speech in order to create a representation, whereas language generation addresses the opposite direction, creating text from representation. The topic at hand is related to the former of the two.

Taking a closer look at the historic origins of NLP besides the domain origins shows that NLP goes all the way back to the 1940s. In these early times, Machine Translation (MT) was explored which is the root of NLP [3, 4]. The first MT descriptions go back to Weaver's article about translation from the year 1949 (later published as a book section in 1955) [10]. In the article, Weaver's thoughts on the possibilities and potential obstacles regarding the translation of languages by machines are outlined. This was arguably influenced by the war circumstances of the subsequent years and insights gained during decryption and enemy message interception, which are concepts that are reflected in Weaver's description as well [3]. As a result, research into MT began based on stochastic and statistical approaches that attempted to tackle issues such as different translations of words, meanings, and ambiguities to name a few. It was soon discovered that the task might be more difficult than anticipated. Similar concerns are already mentioned in Weaver's conversations with Professor Norbert Weiner in 1947 [10].

Following the efforts from the 1940s and early 1950s, Chomsky published the idea of generative grammar in 1957 [11] as part of his "Syntactic Structures". The concept describes grammar as a certain set of rules that result in the constellations and combinations of words forming sentences in a given language. Chomsky breaks from popular theories of the time (e.g. Shannon's communication theory [12]) by saying that the structure of language cannot be addressed with pure statistical or empirical methods [13, 14]. In addition, Chomsky continued to work on aspects related to generative grammar all the way into the 1960s [15] and his work ended up defining what is now considered the rationalist approach in NLP that was prominent until the mid 1980s [13, 16]. Furthermore, the concept is part of what is considered universal grammar that evolved over time with humans [16]. Figure 2 shows an example: the structure of a sentence according to Chomsky's approach and constellation: A sentence is divided into a noun phrase (NP) and an additional verb phrase (VP). The latter also includes the object as a noun phrase with the respective determiner (in this case a definite article with T for "the").

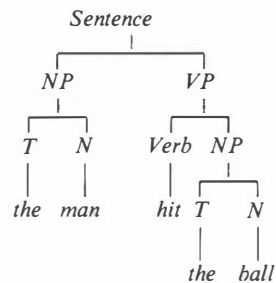

*Fig. 2 - Sentence construct according to generative grammar [11]*

Throughout the 1960s, the movement based on Chomsky's approach of symbolic interpretation and the stochastic/statistical one based on Shannon's methods [12] coexisted and advanced. Noteworthy results of this period were the first parsing systems by Harris [17] as part of the symbolic paradigm. For the stochastically side, the first mentioning of Artificial Intelligence (AI) emerged [16] and Bledsoe and Browning [18] developed the first optical character recognition approach based on the likelihood of each recognized string. Also in the 1960s, Woods published procedural semantics for a question-answering machine [19]. Albeit still based on programmed subroutines, Woods' publications show elements that can be associated with NLU as the answer of a question requires the extraction of semantic meaning from the question. The application was limited, but question-answer machines are still used today in voice assistants, for example. The structure by Woods [19] for the question-answering machine that was initially demonstrated with "what" questions.

During the 1970s and 1980s, the field of NLP grew broader and topics such as NLU emerged, envisioning room beyond NLP in the direction of text/speech recognition and synthesis [20, 21]. NLU was first approached and impressively demonstrated by Winograd [22]. In the publication "Understanding natural language" the authors demonstrate a program that was able to identify and select different shapes and colors in a simulated environment based on given text commands. This work bears strong ties with Woods' work mentioned above [19] and both drove the field of logic-based NLU. Additional noteworthy contributions to this trend include Roger Schank and his colleagues work on language understanding programs [23-25].

In the second half of the 1980s and early 1990s, statistical approaches re-emerged [26] as the primary focus of NLP/NLU, moving away from the symbolic ideas mainly shaped by Chomsky [13, 15]. This popularity of stochastic methods in speech/language processing was significantly driven by IBM's Thomas J. Watson Research Center [16]. The re-emergence came with novel speech-recognition models that sought to bring NLU and speech analysis closer together [27]. Eventually, before the beginning of the twenty-first millennium, The described changes and refocused popularity had made probabilistic models the predominant force in NLP and the rapid increase in computer power as well as the expansion fo the internet created a need for language-based information processing and extraction [16]. The combination of these components and circumstances lead to a more unified but changed field of NLP/NLU and eventually gave way to the rise of Machine Learning in the first decades of the twenty-first century.

In the last 20 years, the interest in NLP has further increased rapidly in conjunction with the stark adoption of Machine Learning [21]. The pace that the subject and topics had picked up by the end of the 1990s was unprecedented [16] especially since the developments in the decades before were described as incremental by experts [28]. As a result, numerous banks and datasets were published in a few years [29-31]. These banks were collections and trees that contained text structures with underlying semantic information and details about syntactics. With the help of such banks and trees, further advances in parsing, tagging, reference resolution [32], and information extraction were also enabled [16]. In addition to the published banks, the ML applications incorporated models such as the Bayesian Analysis [33] and maximum entropy to train systems to process text in accordance with semantic, morphological, and or syntactic parameters [32]. Notable results were significant improvement in various directions, even parts of the aforementioned, such as disambiguation, the answering of questions by a machine and summarization [32]. Until the time of this writing (June 2021) NLP and NLU are active topics of research and computer linguistics in total is described as an active field in AI research [34, 35].

In summary, NLP has gone though various changes over time. It began with Machine Translation and stochastic approaches, then transitioned to semantic and symbolic methods. A broader expansion accompanying the emergence of concepts of NLU and speech recognition enabled regained popularity of stochastic approaches before rapid changes in computer hardware and expansion of the web supercharged the progress of NLP, NLU, speech recognition, and machine translation soon followed and accompanied by ML and AI. Also, potential future developments have been explored and considered as shown in Figure 3 based on the predictions and analyses by Cambria and White [36] who predict a stop of the reliance on word-based techniques to utilize semantics more broadly and effectively:

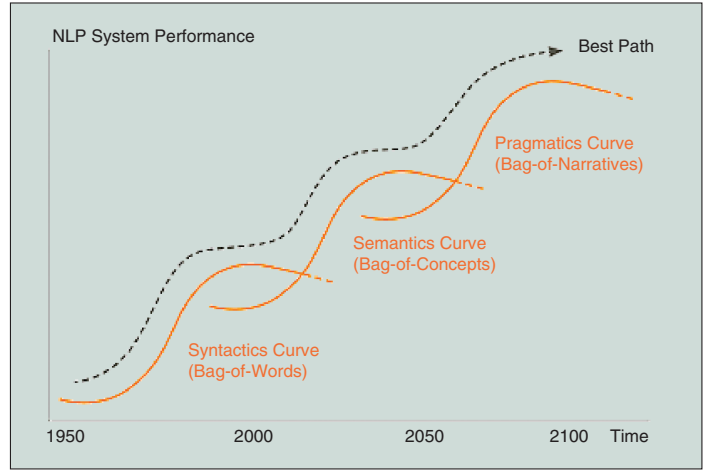

*Fig. 3 - Considered Evolution of NLP over Time [36]*

*Natural Language Processing for Requirements Engineering (NLP4RE)*

Looking at NLP in conjunction with RE, various approaches can be discovered as well with some of them going back to the time that can be considered the mainstream beginning of RE in the late 1990s [37]. This long history and different directions have made the space of what is today called Natural Language Processing for Requirements Engineerings - short NLP4RE - significantly diverse. As a result of this diversity, not all approaches achieve popularity, for example due to their niche existence and or special purpose. Thus, looking for approaches to consider and asses is a difficult task. Albeit, studies have been conducted that target precisely this issue of conflation. The most comprehensive one to date was published by Liping Zhao et al. [37] in April 2021. This study assessed the space of NLP4RE regrading tools, solutions, and techniques. The result were 404 relevant studies that the authors classified and extracted 130 tools from. These tools relied on 66 different methodologies and 25 NLP resources [37]. In addition, Zhao et al. emphasized that most of the tools and techniques have not made it out of laboratory settings, are commonly focused on the analysis of requirements, and require specifications [37, 38], which is in line with the research at hand. Due to its peer-reviews nature, extensiveness, and confirmation by other publications [39, 40] it forms a big part of the sources of the presented work. In addition, the results and insights presented by Zhao et al. also have been confirmed by work of the authors, whose research and discovered results turned out to be overlapping sub-sets of Zhao et al.'s. The final outcome and considered publications of Zhao et al. are depicted in Figure 4 below:

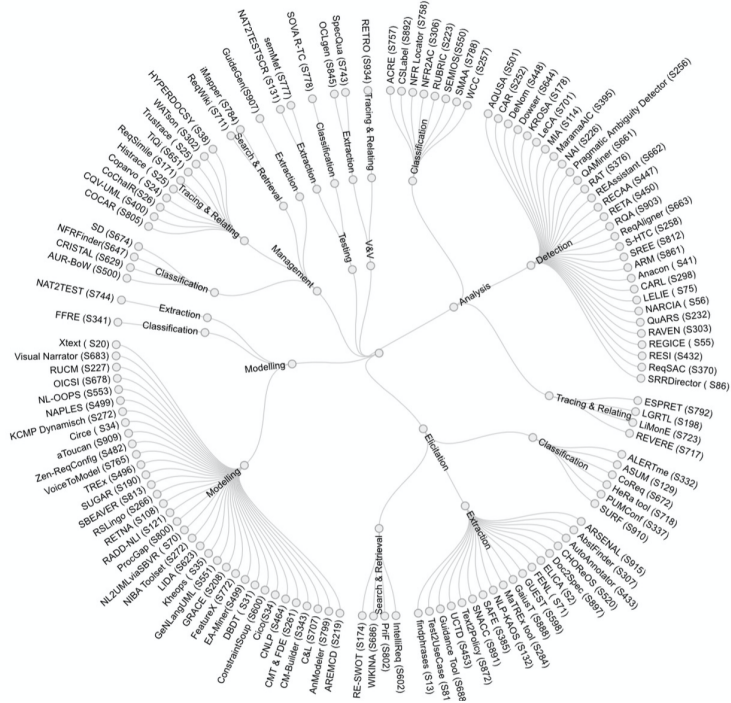

*Fig. 4 - 130 NLP tools discovered and classified by Zhao et al. [37]*

In addition to the publication above, renewed tool/appraoch research had to be added due to the fact that the presented work was conducted after the mentioned publication. This additional round of research also allowed for the identification of currently (as of August 2021) ongoing trends in the field of NLP4RE. As such, the additional publications are outlined hereinafter.

First, Mengyuan et al. presented an approach that utilizes NLP to extract domain models for control systems [41]. Their approach is based on the Rupp's template for requirements and allows for the extraction as well as visualization of models. Second, two tools addressing causality in requirements and the detection thereof were discovered. These two tools, CiRA [42] and CATE [43], address causal relationships within requirements. Such relationships are assessed as to which requirements causes or depend on others. Third, Sonbol, Rebdawi, and Ghneim published their approach called ReqVec that allows for the deduction of semantic relationships as well as classification of requirements [44]. This approach, based on Word2vec, showed a high efficiency in the presented tests and is also considered in the next section. Fourth, Schlutter and Vogelsang published an approach to trace the connections between requirements, which they call "Trace Link Recovery" [45]. This approach is utilizing an explicit content description of the requirements in the form of a semantic relations graph that allowed for the tracing of connections within. Lastly, van Vliet et al. [46] present an approach for NLP based on crowdsourcing to solve shortcomings of what they consider a lack of accuracy and reliability of current approaches. The approach is not strictly related to RE, but nevertheless addresses requirements in their general form and the management thereof.

All in all, the last paragraph shows an active and ongoing research in the field of NLP4RE. Furthermore, the different directions show that there are still various topics and different ideas being pursued. This further supports the purpose and ideas of the presented work, as structure and additional organization is valuable. Also, such organization can contribute to currently identified challenges, as outlined by Kaddari et al. [6].

IV. ANALYSIS AND EVALUATION

With the theoretical foundation and sources presented above, the actual analysis was conducted. As shown in Figure 1, the chosen sources were assessed based on a defined set of criteria to better understand and analyze the space and existing approaches. The definition of said criteria forms the first part of this section and allows the subsequent analysis as an application.

*Criteria and Analysis Foundation*

The criteria to define had to allow the assessment of each approach separately, as well as in a comparative form. This also considers the fact that despite the individual assessment, combinations are possible and planned to be evaluated in a subsequent step as it is possible that the application of two or more tools in conjunction might eliminate certain drawbacks and limitations. To tie in the criteria with the purpose of the research, they were directly derived from the extraction of structure and RE. In addition, the criteria were phrased and chosen to allow for a practical application to exclude options that do not apply due to a lack of implementation possibilities. As such, the following criteria were defined (ranked by importance in descending order):

- **Possibility to elicit structure or topology from source text** - this criteria is the main focus and thus the make or break factor. Approaches that do not allow for the extraction of structure at least with some additional processing are to be considered not suitable.
- **Open-source, acquirable, or accessible via API** - tools that are not accessible or usable for purchase do only serve as concepts and thus have to be considered of less value. Information can still be gained, but not directly applied, which is why a direct application would be preferable.
- **Necessity for full or partial supervision/validation -** Involving or requiring humans for supervision or result verification is, while sometimes necessary, a step that can severely limit the application of an approach and thus has to be optional if it can be avoided. This is also due to the fact that supervision can come with additional limitations and or prerequisites, which should be avoided to not interfere with the limitation criterion below.
- **Proof of concept existing or shown** - a proof of concept is essential to show the capabilities of the tool/approach as a methodology without validation has to be considered conceptual until validated. As such, tools and solutions with proof of concepts or validation are to be preferred.
- **No input limitations regarding format and structure** - input limitations, while sometimes necessary, greatly reduce the applicability of tools and approaches and if the limitations do not align with the input, a solution has to be considered not suitable without adjustment.
- **Modern architecture or active development/support -** The application and actual implementation stands and falls with the compatibility of the architecture and or programming language. If a solution is built on an architecture that is not timely anymore, it has to be considered outdated. The only way an older or uncommon architecture can thus be acceptable, is if proper support and active development is available.
- **Modifiability or possibility for settings adjustment** - settings or modifiability can impact and potentially change the severity of the input limitation criteria, which is why they should be considered also for future adaptation and transfer.

With the criteria, different approaches can be assessed and compared. To allow for a unified evaluation, the possibilities in the table below were set to ensure reproducibility. Naturally, due to the nature of the publications, some criteria were not possible to be assessed since the publications did not include the information necessary. In these cases "UNKNOWN" was used a placeholder. This placeholder does indicate the absence of information, but the assessment of the respective criterion is possible with further information. In addition, each criterion had the possibility to be assigned the answer in the last column to allow for exceptions.

| Criterion | | Possible Answers/Assessments | | | |
|---|---|---|---|---|---|
| A | Possibility to elicit structure | Yes | No | In-Part | Other |
| B | Open-source or acquirable | Open-Source (OS) | Commercial (Comm) | No | Other |
| C | Supervision requirement | Yes | No | In-Part | Other |
| D | Proof of concept | Yes | No | Theoretical | Other |
| E | Input limitations | Yes | No | Other | |
| F | Modern architecture | Yes | No | In-Part | Other |
| G | Modifiability | Yes | No | In-Part | Other |

*Table 1 - Criteria Options and Answers*

*Analysis and Criteria Application*

With the criteria and the rubric outlined in the last section, the actual analysis was conducted. Herein, as shown in Figure 1, the criteria were applied to the set sources. In order to allow for the best comparison between the assessed sources, Table 2 shows the results of all the evaluated approaches in a comprehensive form. To obtain the results, each publication was assessed individually and rated as per Table 1.

With the inclusion of the additions mentioned above, the analysis included 134 publications with 136 tools. For two publications, the presented approach included two tools, but these tools were assessed as one approach and upon further examination, individual evaluation would not have made a difference in the results. With the insights and statistics gained from the analysis, the next section outlines and discusses the exact results of the research.

V. RESULTS AND DISCUSSION

In order to provide the results in a structure way, the following content is divided into three parts: the first outlines the quantitative results of the evaluation, providing statistics and numerical results of the analysis; the second presents specific publications that showed certain features and or attributes worth mentioning and why they stand out; lastly, the third part illustrates interpretations of the former two in form of insights into the statistics and content which also includes the discussion of the results overall.

*Quantitative Results and Statistics*

Summarizing the results of the research, Figure 5 shows the pie charts of the different criteria excluding the "UNKNOWN" answers as in those cases, the publication did not allow for a proper assessment of the respective criteria due to missing information. This result applied to 20 cases of criterion B, two (2) cases of criterion C, two (2) cases of criterion D, three (3) cases of criterion E, and four (4) cases of criterion F. The "Other" category was used for results that did not fit into any of the other answers, but indicated some work in lines with the criterion.

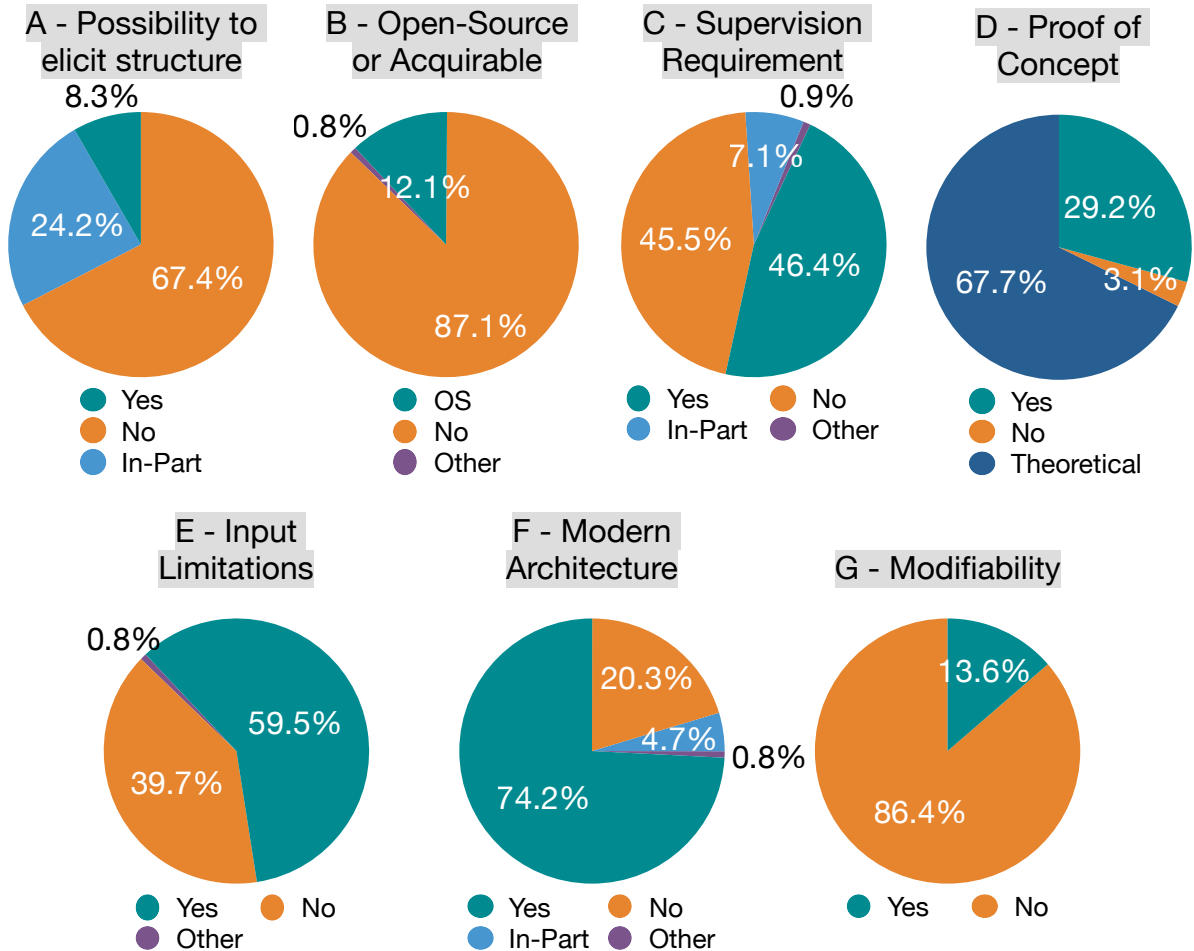

*Fig. 5 - Numeric Evaluation Results*

As the numeric results in Fig. 5 indicate, the most important criteria in the beginning show a significant number of negative or only partially positive results: for example, over 67% of all assessed publications did not show a possibility to extract structure at all. Overall, three (3) publications were identified that did not show any negative results at all [47-49], only partially positive ratings at most. In addition, ten (10) publications were identified that had overall at lest partially positive results except for criteria B and G [44, 50-58], wherein G was in all cases a result of B due to the fact that no access to the source or algorithms does not allow for modification or implementation. The two groups are listed in Table 2 and 3.

| Publication | | | Criteria | | | | | | |
|---|---|---|---|---|---|---|---|---|---|
| | | | A | B | C | D | E | F | G |
| [47] | Ferrari & Gnesi | 2012 | In-Part | OS | No | Theoretical | No | Yes | Yes |
| [48] | Sree-Kumar, Planas, & Clarisó | 2018 | In-Part | OS | No | Theoretical | No | Yes | Yes |
| [49] | Tiwari, Ameta, & Banerjee | 2019 | In-Part | OS | In-Part | Yes | No | Yes | Yes |

*Table 2 - Best rated publications*

| Publication | | | Criteria | | | | | | |
|---|---|---|---|---|---|---|---|---|---|
| | | | A | B | C | D | E | F | G |
| [50] | Al-Safadi | 2009 | In-Part | No | No | Theoretical | No | In-Part | No |
| [51] | Körner & Landhäußer | 2010 | In-Part | No | No | Theoretical | No | In-Part | No |
| [52] | Schulze, Chimiak-Opoka, & Arlow | 2012 | In-Part | No | No | Theoretical | No | Yes | No |
| [53] | Ferrari, Gnesi, & Tolomei | 2013 | Yes | No | No | Theoretical | No | Yes | No |
| [54] | Hamza & Walker | 2015 | Yes | No | No | Theoretical | No | Yes | No |
| [55] | Sannier, et al. | 2017 | Yes | No | No | Theoretical | No | Yes | No |
| [56] | Tahvili et al. | 2018 | Yes | No | No | Yes | No | Yes | No |
| [57] | Li et al. | 2019 | In-Part | No | In-Part | Yes | No | Yes | No |
| [58] | Reinhartz-Berger & Kemelman | 2020 | In-Part | No | No | Theoretical | No | Yes | No |
| [44] | R. Sonbol G. Rebdawi N. Ghneim | 2020 | In-Part | No | No | Theoretical | No | Yes | No |

*Table 3 - Best rated not open-source publications*

As seen above, the three publications in Table 2 show the highest potential albeit none of them allows for full extraction of structure as per criterion A. Nevertheless, these three contenders are assessed in more detail regarding their specific content in the following sub-section.

*Individual Publications to Mention*

The first publication from Table 2 by Ferrari and Gnesi [47] presents a NLP clustering algorithm titled "Sliding Head-Tail Component". This algorithm is supposed to analyze and understand requirement documents from a structural perspective to elicit cluster and relatedness information. In addition, the approach collates the structure of the document itself with the elicited one based on relations. As such, the approach is also described to give recommendations or point to the find mismatches to improve the cohesiveness of the document. Inside the algorithm, the Sliding Head-Tail Component identifies requirements that are lexically related, which are clustered, and also sections that are lexical independent, which allows for partitioning [47]. By differentiating between related and unrelated requirements, the approach by Ferrari and Gnesi extracts what they call "hidden structure" [47], which is the structure resulting from the lexical information instead of the document. To achieve the lexical connections, Ferrari and Gnesi utilized two similarity metrics in conjunction: the first metric is based on the Jaccard index and the second metric is based on the Levenshtein distance [59]. In testing, the approach showed promising results, but was described to also have improvement potential due to false positives [47].

The second well-rated publication, written by Sree-Kumar et al. [48], presents an approach to extract a Feature Model from specification documents. As such, the approach combines various other methods and NLP tools to derive two algorithms as part of the open-source "FeatureX" tool [48]. The first of the two algorithms extracts relationships within the document body and outputs heuristic results. These results are then used in Algorithm 2 to create a feature model candidate. All in all, the work by Sree-Kumar et al. showed promising results regarding the extraction of feature models and also provided improvements over the existing models used as comparison.

Lastly and also the most current contender, is the 2019 publication by Tiwari et al. [49]. In their publication, Tiwari et al. outline an approach that allows for the extraction of use case scenarios from specification and requirement documents. While not the main focus of the approach, structure is extracted within the process as the input is handled by various NLP tools, such as POS taggers and Type Dependency Parsers that feed a rule-based engine (based on heuristics [60]) that allows for the detection of use case names, actors, dependencies, basic flow steps, alternative flows, and post-conditions. Albeit the approach additionally includes the extraction of actual use cases, which involves a human, the above mentioned results of the rule based engine are reminiscent of structure and could be used as such. Yet, the main purpose of the work presented by Taiwan et al. was use case extraction and analysis, of which the structure of the input can be merely seen as a byproduct.

*Interpretation, Insights, and Discussion*

The above presented results and individual publications provide an overview regarding the space assessed by the literature review. Yet, the statistical results and individual contenders, while relevant, only show the surface of the results. Therefore, this sub-section addresses the insights and inferences based on the results.

To begin with, the results shown in the pie charts in Fig. 5 show a significant amount of negative results for all of the top three criteria. As such, it is apparent that most of the assessed approaches do not target the extraction of structure at all, and some only in part. Only 11 publications directly approach the extraction of structure, out of which none were open source. This is in part due to the fact that RE in general has many different areas and topics that NLP4RE can address, such as classification, elicitation, or change management. Thus, structure extraction only forms one sub-task that suffered from divided attention.

In addition, most publications were closed source and not available at all, which for one, makes assessment and evaluation difficult and two, completely prohibits use and modification without difficult and potentially erroneous reverse engineering. The fact that this applies to 115 of the 132 contenders renders this group only useful for concept adaptation and potential transfer/recreation. This fact is also directly related and causes the high negative ratings for criterion G as not accessible algorithms cannot be modified or checked for the possibility to do so. Another relation of criterion B exists with criterion F, as older architecture that is not open source will be significantly more difficult to use or adapt. Fortunately, almost 80% of the assessed publications were of modern architecture or at least in part.

Criterion B, the supervision requirement, together with criterion E, the input limitations, show another characteristic of the assessed research: most a significant amount of approaches (in case of criterion C almost half) comes with limitations either on the input or the process side. This means that automation of the entire approach will be impossible before these limitations or required input can be solved or substituted. Yet, this is not per se a limitation for the extraction of structure, as the conclusion below elaborates.

Lastly, the criterion for proof of concept showed that there is a very low number of actual case studies that are being used for validation purposes. Since over 1/3 of all the assessed contenders showed only theoretical proofs of concept, the validity and real-world applicability has to be regarded es relative. Thus, even if an approach with a theoretical proof of concept were to be adapted or chosen, its correctness would require additional validation.

Adding to the statistical results, the predominately negative ratings of the most important criteria get exacerbated by the fact that, while the criteria individually might have shown some positive results, the overall and thus end results is even smaller. This can be seen in Table 2 and 3, which include only 3 and 10 publications respectively. This means that only 2.2% of all assessed publications were relevant at all based on the chosen criteria and only 7.6% were potentially applicable, but not openly accessible. Hence, less than 10% of the assessed literature showed relevance. To illustrate the effect that the overlap has, Fig. 6 shows the sets of the most significant criteria.

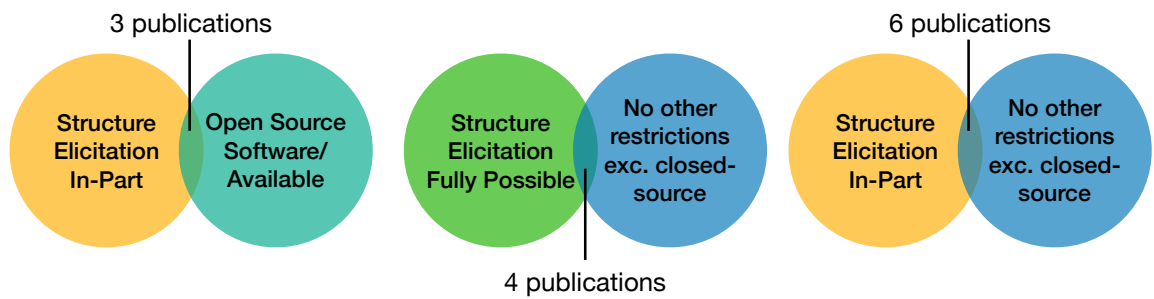

*Fig. 6 - Sets and Overlaps*

As Fig. 6 shows, the overlap of the different sections is under 5% of the total number of contenders for each combination. Thus, it can be deduced that not many publications at all pose potential solutions in a comprehensive manner. As a result, adaptations or additions are essential and cannot be avoided if a solution to the problem outlined in the beginning is to be attempted. Even the best contenders have limitations as outlined hereinafter.

Despite three publications fulfilling all criteria at least in part, none of the them targeted the extraction of requirement structure primarily: the first one mentioned, by Ferrari and Gnesi [47], while allowing for the extraction of structure, mainly focused on the clustering and not the overall structure of the entire input in and by itself; the second contender, by Sree-Kumar et al. [48], targeted the feature model as its primary objective and thus, structure was only a byproduct; lastly, Tiwari et al. [49] addressed the extraction of use cases which included the elicitation of structural elements, but did not address structure comprehensively. Thus, even the top contenders have drawbacks.

This concludes the discussion of the literature review and statistic results. To also outline how the obtained results will be utilized, the next and last section provides a conclusion and outlook for the presented work.

## VI. Conclusion, Limitations, and Outlook

The work presented above shows the results and analysis of the current literature of Natural Language Processing for Requirements Engineering to assess the possibility of structure extraction. 136 tools and approaches to apply NLP to Requirements Engineering were assessed out of which 130 were previously gathered by Zhao et al. [37] and 6 additional ones were added to account for the most recent developments and trends.

Based on the criteria that were derived from the requirement to elicit structure with an applicable approach, all publications and tools were assessed to define their suitability as part of the literature review. Herein, seven criteria were chosen to check for the possibility to utilize or potentially transform existing approaches (see Table 1). The result was a comprehensive assessment database that was analyzed numerically as well as compressively to derive non-numeric insights.

The main insights resulting from the analysis were that no approach completely fulfilled the defined criteria and could thus be deemed applicable. The only contenders that fulfilled the most criteria only targeted the extraction of structure in part and thus require further adaptation to be considered useful. Other approaches and tools that did target the elicitation of structure primarily turned out to be not openly accessible as far as their code basis is concerned, which makes them only useful conceptually and as such requires reconstruction of the actual tool. All in all, the analysis has shown that no approach exists that fulfills or addresses the problem and situation set forth in the introduction and thus, the creation of a novel tool to tackle this task is auspicious.

In addition to the confirmation of the purpose of the task introduced, further insights were gained from the dataset and literature. For one, the statistics shows a high reliance of the tools on supervision and input requirements, which has to be considered in the work moving forward. Such a wide-spread reliance on supervision and input limitations could be due to general difficulties with unrestricted and universal approaches, which might indicate that initial limitation or even general restriction might have to be considered. A possible solution to this could be the separation of the approach from the domain for example, which would allow for a flexible and modular application. Second, the data and sources also showed that concept proofs and validations are often (over two-thirds) only conducted in a theoretical manner, which, while validating the approach, has to be seen as another limitation due to the fact that actual case studies might show additional complications to consider. Therefore, the real-world application and its potential problems have to also be considered in the development moving forward.

Despite the clear outcome, the work presented underlies certain limitations and requires the potential subsequent steps in the next paragraph. For one, the evaluated sources and data base cannot be seen as entirely complete since the field was restricted to the subset of NLP4RE. As such, it is possible that a solution to the problem presented exists in the space of NLP or elsewhere, which has not been part of the research yet due to the references. Yet, the existence of a solution from another field or NLP space has to be seen with the caveat that requirements and documents thereof are special in many ways and differ from regular text corpora, which would limit the application of an approach that was not initially geared towards RE. Another limitation of the work presented is that the criteria were chosen based on the goals and ideas outlined in the introduction, namely the automatic extraction of semantic structure from requirement documents. Therefore, it is possible that a solution that was ruled out based on the criteria applied would be sufficient and applicable to other use cases of structural extraction. Hence, the transferability of the results has to be checked as to the alignment with the criteria.

All in all, the presented work shows the opportunity for an approach to elicit structure from requirements and as such, the future plans are to develop a tool to utilize for that purpose. This will be conducted in accordance with Fig. 1 to utilize the insights and results of the provided literature review. In addition, other contenders that were not open source (see Table 3) have to be assessed for the contribution in a conceptual form despite the lack of source code access. These conceptual aspects have to be considered in the approach to develop as well.

Lastly, the thoughts and insights outlined in the work at hand also allow for potential inclusion and bilateral benefits when combined with other tools, such as machine learning. One possible scenario could for example be the iterative and machine supported generation of semantic foundations based on documents and data supplied from a source field. This would allow for a consistent and easy to update foundation of the input component. Such options are being assessed and could be also included in the approach to developed.

BIBLIOGRAPHY

[1]  C. Lipizzi, D. Borrelli, and F. Capela, "The "Room Theory": a computational model to account subjectivity into Natural Language Processing," arXiv.org, 2020, arXiv:2005.06059.
[2]  D. M. Berry, E. Kamsties, M. M. Krieger, W. L. S. Lee, and W. L. S. Tran, "From Contract Drafting to Software Specification: Linguistic Sources of Ambiguity," 2003.
[3]  E. D. Liddy, "Natural Language Processing," in *Encyclopedia of Library and Information Science*, 2nd ed.: NY. Marcel Decker, Inc., 2001.
[4]  G. G. Chowdhury, "Natural language processing," *Annual Review of Information Science and Technology,* vol. 37, no. 1, pp. 51-89, 2003, doi: https://doi.org/10.1002/aris.1440370103.
[5]  T. Beysolow, "What Is Natural Language Processing?," in Applied Natural Language Processing with Python : Implementing Machine Learning and Deep Learning Algorithms for Natural Language Processing. Berkeley, CA: Apress, 2018, pp. 1-12.
[6]  Z. Kaddari, Y. Mellah, J. Berrich, M. G. Belkasmi, and T. Bouchentouf, "Natural Language Processing: Challenges and Future Directions," Cham, 2021: Springer International Publishing, in Artificial Intelligence and Industrial Applications, pp. 236-246.
[7]  J. Davis, K. Mengersen, S. Bennett, and L. Mazerolle, "Viewing systematic reviews and meta-analysis in social research through different lenses," *SpringerPlus,* vol. 3, no. 1, p. 511, 2014, doi: 10.1186/2193-1801-3-511.
[8]  R. J. Torraco, "Writing Integrative Literature Reviews: Guidelines and Examples," *Human Resource Development Review,* vol. 4, no. 3, pp. 356-367, 2005, doi: 10.1177/1534484305278283.
[9]  T. Mazumdar, S. P. Raj, and I. Sinha, "Reference Price Research: Review and Propositions," *Journal of Marketing,* vol. 69, no. 4, pp. 84-102, 2005, doi: 10.1509/jmkg.2005.69.4.84.
[10] W. Weaver, "Translation," in *Machine Translation of Languages*, W. N. Locke and A. D. Booth Eds. Cambridge, MA: MIT Press, 1955.
[11] N. Chomsky, *Syntactic Structures*. Mouton & Co., 1957.
[12] C. E. Shannon, "A mathematical theory of communication," *The Bell System Technical Journal,* vol. 27, no. 3, pp. 379-423, 1948, doi: 10.1002/j.1538-7305.1948.tb01338.x.
[13] C. Manning and H. Schütze, Foundations of Statistical Natural Language Processing. Cambridge, MA: MIT Press, 1999.
[14] D. A. Dahl, "Natural Language Processing: Past, Present and Future," in *Mobile Speech and Advanced Natural Language Solutions*, A. Neustein and J. A. Markowitz Eds. New York, NY: Springer New York, 2013, pp. 49-73.
[15] N. Chomsky, *Aspects of the Theory of Syntax*. Cambridge, MA: MIT Press, 1965.
[16] D. Jurafsky and J. H. Martin, *Speech and Language Processing*, 2nd ed. Upper Saddle River, NJ: Prentice Hall, 2008.
[17] Z. S. Harris, *String Analysis Of Sentence Structure*. The Hauge, Netherlands: Mouton Publishers, 1962.
[18] W. W. Bledsoe and I. Browning, "Pattern recognition and reading by machine," presented at the Papers presented at the December 1-3, 1959, eastern joint IRE-AIEE-ACM computer conference, Boston, Massachusetts, 1959. [Online]. Available: https://doi.org/10.1145/1460299.1460326.
[19] W. A. Woods, "Procedural semantics for a question-answering machine," in *Proceedings of the December 9-11, 1968, fall joint computer conference, part I*, San Francisco, California, 1968: Association for Computing Machinery, pp. 457–471, doi: 10.1145/1476589.1476653.
[20] L. Rabiner and B.-H. Juang, *Fundamentals of Speech Recognition*. Englewood Cliffs, NJ: Prentice Hall, 1993.
[21] T. Beysolow II, "What Is Natural Language Processing?," in Applied Natural Language Processing with Python : Implementing Machine Learning and Deep Learning Algorithms for Natural Language Processing. Berkeley, CA: Apress, 2018, pp. 1-12.
[22] T. Winograd, "Understanding natural language," *Cognitive Psychology,* vol. 3, no. 1, pp. 1-191, 1972, doi: 10.1016/0010-0285(72)90002-3.
[23] R. C. Schank, "Conceptual dependency: A theory of natural language understanding," *Cognitive Psychology,* vol. 3, no. 4, pp. 552-631, 1972, doi: 10.1016/0010-0285(72)90022-9.
[24] R. C. Schank and R. P. Abelson, *Scripts, plans, goals and understanding: An inquiry into human knowledge structures* (Scripts, plans, goals and understanding: An inquiry into human knowledge structures.). Oxford, England: Lawrence Erlbaum, 1977.
[25] R. C. Schank and C. K. Riesbeck, *Inside Computer Understanding*. New York, NY: Psychology Press, 1981.
[26] K.-F. Lee and R. Reddy, Automatic Speech Recognition: The Development of the Sphinx Recognition System. New York, NY: Springer Science+Busienss Media, 1988.
[27] L. Hirshman, "Overview of the DARPA Speech and Natural Language Workshop," presented at the Proceedings of the workshop on Speech and Natural Language, Philadelphia, PA, 1989.
[28] Futures Group Glastonbury CT, "State of the Art of Natural Language Processing," Defense Technical Information Center, ADA188112, 1987.
[29] M. Marcus, B. Santorini, and M. A. Marcinkiewicz, "Building a Large Annotated Corpus of English: The Penn Treebank," University of Pennsylvania Department of Computer and Information Science, MS- CIS-93-87, 1993.
[30] J. Pustejovsky et al., "The TIMEBANK Corpus," in Proceedings of the Corpus Linguistics Conference, 2003, pp. 647–656.

[31] M. Palmer, D. Gildea, and P. Kingsbury, "The Proposition Bank: An Annotated Corpus of Semantic Roles," *Computational Linguistics,* vol. 31, no. 1, pp. 71-106, 2005, doi: 10.1162/0891201053630264.
[32] R. Kibble, *Introduction to natural language processing*. London, United Kingdom: University of London, 2013.
[33] A. Gelman, J. B. Carlin, H. S. Stern, D. B. Dunson, A. Vehtari, and D. B. Rubin, *Bayesian Data Analysis*, 3rd ed. Boca Raton, FL: CRC Press, 2013.
[34] T. Young, D. Hazarika, S. Poria, and E. Cambria, "Recent Trends in Deep Learning Based Natural Language Processing [Review Article]," *IEEE Computational Intelligence Magazine,* vol. 13, no. 3, pp. 55-75, 2018, doi: 10.1109/MCI.2018.2840738.
[35] E. Ghazizadeh and P. Zhu, "A Systematic Literature Review of Natural Language Processing: Current State, Challenges and Risks," in *Proceedings of the Future Technologies Conference (FTC) 2020, Volume 1*, Cham, K. Arai, S. Kapoor, and R. Bhatia, Eds., 2021// 2021: Springer International Publishing, pp. 634-647.
[36] E. Cambria and B. White, "Jumping NLP Curves: A Review of Natural Language Processing Research [Review Article]," *IEEE Computational Intelligence Magazine,* vol. 9, no. 2, pp. 48-57, 2014, doi: 10.1109/MCI.2014.2307227.
[37] L. Zhao *et al.*, "Natural Language Processing for Requirements Engineering: A Systematic Mapping Study," *ACM Comput. Surv.,* vol. 54, no. 3, p. Article 55, 2021, doi: 10.1145/3444689.
[38] A. Ferrari, L. Zhao, and W. Alhoshan, "NLP for Requirements Engineering: Tasks, Techniques, Tools, and Technologies," in *2021 IEEE/ACM 43rd International Conference on Software Engineering: Companion Proceedings (ICSE-Companion)*, 25-28 May 2021 2021, pp. 322-323, doi: 10.1109/ICSE-Companion52605.2021.00137.
[39] A. Alzayed and A. Al-Hunaiyyan, "A Bird's Eye View of Natural Language Processing and Requirements Engineering," *International Journal of Advanced Computer Science and Applications,* vol. 12, no. 5, pp. 81-90, 2021, doi: 10.14569/IJACSA.2021.0120512.
[40] H. Schrieber, M. Anders, B. Paech, and K. Schneider, "A Vision of Understanding the Users' View on Software," in *Joint Proceedings of REFSQ-2021 Workshops, OpenRE, Posters and Tools Track, and Doctoral Symposium*, Essen, Germany, F. B. Aydemir *et al.*, Eds., 2021.
[41] Y. Mengyuan *et al.*, "Automatic Generation Method of Airborne Display and Control System Requirement Domain Model Based on NLP," in *2021 IEEE 6th International Conference on Computer and Communication Systems (ICCCS)*, 23-26 April 2021 2021, pp. 1042-1046, doi: 10.1109/ICCCS52626.2021.9449277.
[42] J. Fischbach *et al.*, "Automatic Detection of Causality in Requirement Artifacts: The CiRA Approach," Cham, 2021: Springer International Publishing, in Requirements Engineering: Foundation for Software Quality, pp. 19-36, doi: 10.1007/978-3-030-73128-1_2.
[43] N. Jadallah, J. Fischbach, J. Frattini, and A. Vogelsang, "CATE: CAusality Tree Extractor from Natural Language Requirements," arXiv.org, 2021, arXiv:2107.10023.
[44] R. Sonbol, G. Rebdawi, and N. Ghneim, "Towards a Semantic Representation for Functional Software Requirements," in *2020 IEEE Seventh International Workshop on Artificial Intelligence for Requirements Engineering (AIRE)*, 1-1 Sept. 2020 2020, pp. 1-8, doi: 10.1109/AIRE51212.2020.00007.
[45] A. Schlutter and A. Vogelsang, "Trace Link Recovery using Semantic Relation Graphs and Spreading Activation," in *2020 IEEE 28th International Requirements Engineering Conference (RE)*, 31 Aug.-4 Sept. 2020 2020, pp. 20-31, doi: 10.1109/RE48521.2020.00015.
[46] M. van Vliet, E. C. Groen, F. Dalpiaz, and S. Brinkkemper, "Identifying and Classifying User Requirements in Online Feedback via Crowdsourcing," Cham, 2020: Springer International Publishing, in Requirements Engineering: Foundation for Software Quality, pp. 143-159.
[47] A. Ferrari and S. Gnesi, "Using collective intelligence to detect pragmatic ambiguities," in *2012 20th IEEE International Requirements Engineering Conference (RE)*, 24-28 Sept. 2012 2012, pp. 191-200, doi: 10.1109/RE.2012.6345803.
[48] A. Sree-Kumar, E. Planas, and R. Clarisó, "Extracting software product line feature models from natural language specifications," presented at the Proceedings of the 22nd International Systems and Software Product Line Conference - Volume 1, Gothenburg, Sweden, 2018. [Online]. Available: https://doi.org/10.1145/3233027.3233029.
[49] S. Tiwari, D. Ameta, and A. Banerjee, "An Approach to Identify Use Case Scenarios from Textual Requirements Specification," presented at the Proceedings of the 12th Innovations on Software Engineering Conference (formerly known as India Software Engineering Conference), Pune, India, 2019. [Online]. Available: https://doi.org/10.1145/3299771.3299774.
[50] L. A. E. Al-Safadi, "Natural Language Processing for Conceptual Modeling," *International Journal of Digital Content Technology and its Applications,* vol. 3, no. 3, pp. 47-59, 2009, doi: 10.4156/jdcta.vol3.issue3.6.
[51] S. J. Körner and M. Landhäußer, "Semantic Enriching of Natural Language Texts with Automatic Thematic Role Annotation," Berlin, Heidelberg, 2010: Springer Berlin Heidelberg, in Natural Language Processing and Information Systems, pp. 92-99.
[52] G. Schulze, J. Chimiak-Opoka, and J. Arlow, "An Approach for Synchronizing UML Models and Narrative Text in Literate Modeling," Berlin, Heidelberg, 2012: Springer Berlin Heidelberg, in Model Driven Engineering Languages and Systems, pp. 595-608.
[53] A. Ferrari, S. Gnesi, and G. Tolomei, "Using Clustering to Improve the Structure of Natural Language Requirements Documents," Berlin, Heidelberg, 2013: Springer Berlin Heidelberg, in Requirements Engineering: Foundation for Software Quality, pp. 34-49.
[54] M. Hamza and R. J. Walker, "Recommending Features and Feature Relationships from Requirements Documents for Software Product Lines," in *2015 IEEE/ACM 4th International Workshop on Realizing Artificial Intelligence Synergies in Software Engineering*, 17-17 May 2015 2015, pp. 25-31, doi: 10.1109/RAISE.2015.12.
[55] N. Sannier, M. Adedjouma, M. Sabetzadeh, and L. Briand, "An automated framework for detection and resolution of cross references in legal texts," *Requirements Engineering,* vol. 22, no. 2, pp. 215-237, 2017, doi: 10.1007/s00766-015-0241-3.
[56] S. Tahvili et al., "Functional Dependency Detection for Integration Test Cases," in 2018 IEEE International Conference on Software Quality, Reliability and Security Companion (QRS-C), 16-20 July 2018 2018, pp. 207-214, doi: 10.1109/QRS-C.2018.00047.
[57] Y. Li, T. Yue, S. Ali, and L. Zhang, "Enabling automated requirements reuse and configuration," *Software & Systems Modeling,* vol. 18, no. 3, pp. 2177-2211, 2019, doi: 10.1007/s10270-017-0641-6.
[58] I. Reinhartz-Berger and M. Kemelman, "Extracting core requirements for software product lines," *Requirements Engineering,* vol. 25, no. 1, pp. 47-65, 2020, doi: 10.1007/s00766-018-0307-0.
[59] V. I. Levenshtein, "Binary codes capable of correcting deletions, insertions, and reversals," *Soviet Physics Doklady,* vol. 10, no. 8, pp. 707-710, 1966.
[60] S. Tiwari and A. Gupta, "A systematic literature review of use case specifications research," *Information and Software Technology,* vol. 67, pp. 128-158, 2015, doi: 10.1016/j.infsof.2015.06.004.